\documentclass[sigconf]{aamas} 
\usepackage{balance} % for balancing columns on the final page
\usepackage{algpseudocode}
\usepackage{algorithm} 
\usepackage{stfloats} % 
\usepackage{subfigure}
\usepackage{enumitem}
\usepackage{balance} % for balancing columns on the final page

\setcopyright{ifaamas}
\acmConference[AAMAS '23]{Proc.\@ of the 22nd International Conference
on Autonomous Agents and Multiagent Systems (AAMAS 2023)}{May 29 -- June 2, 2023}
{London, United Kingdom}{A.~Ricci, W.~Yeoh, N.~Agmon, B.~An (eds.)}
\copyrightyear{2023}
\acmYear{2023}
\acmDOI{}
\acmPrice{}
\acmISBN{}

\acmSubmissionID{fp0974}

\title[AAMAS-2023 Formatting Instructions]{MADDM: Multi-Advisor Dynamic Binary Decision-Making by Maximizing the Utility}
\author{Zhaori Guo}
\affiliation{
  \institution{University of Southampton}
  \country{United Kingdom}}
\email{zg2n19@soton.ac.uk}

\author{Timothy J. Norman}
\affiliation{
  \institution{University of Southampton}
  \country{United Kingdom}}
\email{t.j.norman@soton.ac.uk}

\author{Enrico H. Gerding}
\affiliation{
  \institution{University of Southampton}
  \country{United Kingdom}}
\email{eg@ecs.soton.ac.uk}

\begin{abstract}
Being able to infer ground truth from the responses of multiple imperfect advisors is a problem of crucial importance in many decision-making applications, such as lending, trading, investment, and crowd-sourcing.
In practice, however, gathering answers from a set of advisors has a cost. 
Therefore, finding an advisor selection strategy that retrieves a reliable answer and maximizes the overall utility is a challenging problem.  
To address this problem, we  propose a novel strategy for optimally selecting a set of advisers in a sequential binary decision-making setting, where multiple decisions need to be  made over time. Crucially, we assume no access to ground truth and no prior knowledge about the reliability of advisers. Specifically, our approach considers how to simultaneously (1) select advisors by balancing the advisors' costs and the value of making correct decisions, (2) learn the trustworthiness of advisers dynamically without prior information by asking multiple advisers, and (3) make optimal decisions without access to the ground truth, improving this over time. We evaluate our algorithm through several numerical experiments. The results show that our approach outperforms two other methods that combine state-of-the-art models.
To address this problem, we  propose a novel strategy for optimally selecting a set of advisers in a sequential binary decision-making setting, where multiple decisions need to be  made over time. Crucially, we assume no access to ground truth and no prior knowledge about the reliability of advisers. Specifically, our approach considers how to simultaneously (1) select advisors by balancing the advisors' costs and the value of making correct decisions, (2) learn the trustworthiness of advisers dynamically without prior information by asking multiple advisers, and (3) make optimal decisions without access to the ground truth, improving this over time. We evaluate our algorithm through several numerical experiments. The results show that our approach outperforms two other methods that combine state-of-the-art models.

\end{abstract}
\keywords{Trust and Reputation; Crowdsourcing; Truth Inference}

\newcommand{\BibTeX}{\rm B\kern-.05em{\sc i\kern-.025em b}\kern-.08em\TeX}

\begin{document}

\pagestyle{fancy}
\fancyhead{}

%%% The next command prints the information defined in the preamble.

\maketitle 
%%%%%%%%%%%%%%%%%%%%%%%%%%%%%%%%%%%%%%%%%%%%%%%%%%%%%%%%%%%%%%%%%%%%%%%%

\section{Introduction}
Many situations rely on expert advice to make decisions, and often there is no objectively correct answer. Examples are wide-ranging and include crowdsourcing, machine learning ensemble models, or loan approvals. In such settings, and following the principles of the wisdom of the crowd \cite{landemore2012collective, zheng2017truth}, it may be better to rely on the expertise of multiple advisers, especially if the stakes are high. However, asking for multiple advisers comes at a cost, and the reliability or \emph{quality} of advisers can differ. Therefore,  knowing how many and who to ask is a challenging task. In addition, typically, multiple sequential decisions need to be made, and the reliability of individual advisers can be learned over time. A good strategy for doing so is not obvious, however, when ground truth is not available. To address these challenges, we design a novel method for maximizing the utility in sequential, binary, multi-advisor decision-making problems for settings with no ground truth. 

These types of scenarios are extensively studied in various fields where dynamic pricing for advisors is considered \cite{tong2018dynamic, wang2018real, miao2022dynamically}. 
Tong \textit{et al.} \cite{tong2018dynamic} focus on pricing the advisors in different regions and decided by the relationship between supply and demand in spatial crowdsourcing tasks. They do not consider advisors having different qualities, however. 
Miao \textit{et al.} \cite{miao2022dynamically} and Wang \textit{et al.} \cite{wang2018real} also assume advisors cost the same but give additional rewards to advisors with more contributions. However, when there is no real-time feedback on the ground truth, it is difficult to determine who should get additional rewards. Therefore, the same price for advisors with different qualities is unrealistic; in contrast, our work considers advisors with different qualities and prices. 

Other research considers a fixed budget constraint \cite{tran2012knapsack, tran2014efficient, xia2015thompson, zhou2018budget, cayci2020budget}. However, these papers do not consider that decisions might have different values and costs associated with getting them wrong. Consider the following lending decision scenario. If a $\$$1,000 loan at 9$\%$ interest is repaid, it will make a $\$$90 profit, but it can result in a loss of $\$$1,000 if the borrower defaults. Such high-risk decisions require a more reliable assessment, potentially requiring multiple costly advisers, whereas low-value, low-risk decisions may only need a single one. Therefore, we should consider selecting a group of advisors with different qualities and prices to balance potential profits and risks associated with a decision.

Research grounded in multi-armed bandit methods is also relevant here \cite{kurenkov2019ac, tran2014efficient, tran2012knapsack, xia2015thompson}. However, these works assume that  that the ground truth is available following every decision, which means that advisors' trustworthiness can be reliably updated. Instead, our work infers the reliability of the answer by the decision model, thereby avoiding the need for ground truth. Assessing the reliability of the answer can help us give reasonable update evidence for building models of the trustworthiness of advisors. To summarise, Table \ref{tab:compare} provides an overview of the state of the art and how it compares to the problem we are addressing.  

In more detail, we present a novel method, called Multi-Advisor Dynamic Decision-Making Method (MADDM), to address the limitations of existing approaches described above. MADDM (see Section \ref{Sec: MTIRL} for details) integrates and extends several state-of-the-art methods and consists of three interdependent components: trust assessment, advisor selection, and decision-making. Trust assessment builds and maintains models of the trustworthiness of each advisor. For every sequential decision, advisor selection identifies which advisors to consult. This is similar to a multi-armed bandit problem, which requires a balance of exploration and exploitation. We use Thompson Sampling combined with the decision-making model to compute each advisor's expected marginal contribution and select advisers until the marginal contribution is negative. The third component uses the set of answers from the advisor selected to make a decision using the \textit{Bayesian Weighted Voting Ensemble} (BWVE) method proposed in \cite{zhaori2022}. 
In addition, we conduct extensive experiments (Section \ref{Sec: mtexper}) that compare MADDM to a variety of methods that combine state-of-the-art approaches, including budget-limited decision making, $\epsilon$-greedy selection, and expectation maximization, and we benchmark performance against the optimal utility that could be gained with perfect knowledge. The results show that MADDM outperforms the other two methods in almost all environments. 

Before presenting MADDM in detail, in what follows we first formalize our problem domain.

\begin{table*}[ht]
\centering
\caption{Comparison of MADDM with state of the art. TS = Thompson Sampling; MV = Majority Voting; BWVE = Bayesian Weighted Voting Ensemble; EM = Expectation-Maximization.}\label{tab:compare}
\begin{tabular}{rllllllll}
\hline
Setting & $\epsilon$-First \cite{tran2014efficient}& ZenCrowd \cite{demartini2012zencrowd}& SBB \cite{josang2016generalising}& ACT \cite{kurenkov2019ac} & DEMV \cite{tao2021differential} & BAL \cite{gemalmaz2021accounting}& MTIRL \citep{zhaori2022} & MADDM\\
\hline
    sequential &\checkmark & $\times$     & \checkmark     & \checkmark     & $\times$     & $\times$     & \checkmark     & \checkmark \\
    truth inference& $\times$     & \checkmark     & $\times$     & $\times$     & \checkmark     & \checkmark     & \checkmark     & \checkmark \\
    multi-advisor for one task & $\times$     & \checkmark     & $\times$     & $\times$     & \checkmark     & \checkmark     & \checkmark     & \checkmark \\
    budget-limited & \checkmark     & $\times$     & $\times$     & $\times$     & $\times$     & $\times$     & $\times$     & \checkmark \\
    different task values & $\times$     & $\times$     & $\times$     & $\times$     & $\times$     & $\times$     & $\times$     & \checkmark \\
    different advisor's price & \checkmark     & $\times$     & $\times$     & $\times$     & $\times$     & $\times$     & $\times$     & \checkmark \\
    trustworthiness assessment & \checkmark     & \checkmark     & \checkmark     & \checkmark     & $\times$     & \checkmark     & \checkmark     & \checkmark \\
    insufficient samples & \checkmark     & $\times$     & $\times$     & \checkmark    & $\times$      & $\times$     & \checkmark     & \checkmark \\
    aggregation method & $\times$     & EM    & Bayesian & $\times$     & MV    & EM    & BWVE  & BWVE \\
    advisor selection & $\epsilon$-first & $\times$     & $\times$     & TS    & $\times$     & $\times$     & $\times$     & TS \\
\hline
\end{tabular}

\end{table*}

\section{Problem Formalization} \label{Sec: modelform}
 Let $D$ be the set of decisions and $X$ be a set of advisors.
For every decision $d \in D$, the decision-maker needs to choose a unique answer with a binary value, namely $a_{d} \in \{-1,1\}$.
For simplicity but without loss of generality, we assume that the correct value, i.e. the ground truth, denoted by $a^*_d$,  is positive, i.e.  $a^*_d=1$. 
Given a decision $d$, $v^+_{d}$ is the value that the decision-maker gets if the answer is correct.
We denote with $v^-_{d}$ the value that the decision-maker pays if the answer it infers is wrong. 
Therefore, the value of the decision is represented by the tuple $v^\pm_d = (v^+_d,v^-_d)$.
Moreover, since we rely on advisors to answer queries to inform decisions, we need to incentive them by introducing a payment system. 
For each advisor $x \in X$, $c_x$ is its price. For any given $d\in D$, the decision-maker must select a subset of advisors $Y_d\subseteq X$.

The choice of advisors also depends on their trustworthiness.
For each advisor, $x\in X$, $\tau_x$ is its trustworthiness, which is updated after every decision for which that advisor is consulted.
Finally, we denote with $\vec c$ and $\vec \tau$ the vectors containing all the advisors' prices and trustworthiness values, respectively. 
We, therefore, describe any possible selection through a function $s$ that, to every tuple $I := (d, \vec\tau, v^\pm_d, \vec c)\in D\times [0,1]^{\mid X \mid}\times [0,+\infty]^2\times [0,+\infty]^{\mid X \mid}$, associates a subset of advisors $Y_d\in \mathcal{P}(X)$, where $|X|$ is the cardinality of $X$ and $\mathcal{P}(X)$ is the power set of $X$; we call $s$ the \textit{selection function}. 
Table~\ref{tab:variables_and_parameter} gives an overview of the main variables and parameters used.

For any given $d\in D$, we denote with $P_{d,s} \subseteq s(I) \subseteq X$ the set of advisors who give positive answers to decision $d$. Similarly, we denote with $N_{d,s}\subseteq s(I) \subseteq X$ the set of advisors who give a negative answer to decision $d$. When it is clear from the context, we simplify the notation and use $P_d$ and $N_d$ over $P_{d,s}$ and $N_{d,s}$, respectively. Note that $P_{d,s} \cap N_{d,s} = \emptyset$ and  $P_{d,s} \cup N_{d,s} = s(I)$ for every $d\in D$.

We assume that, for any given decision, $d$, there exists a true answer $a^*_d$, but this is never revealed to the decision maker. 
Therefore, we use $a_d = f(P_d, N_d)$ to refer to the decision-making function of our inference model. This is a function of the advisors' responses in $P_d$ and  $N_d$. Let $v_d \in {v^+_d,v^-_d}$ denote the value that the decision-maker gets from the decision $d$, and let $a_d^*$ denote the ground truth of the decision.
If $a_d=a_d^*$, we say that the answer is correct and $v_d = v^+_d$. Otherwise, we say that the answer is wrong and $v_d = v^-_d$.
\begin{table}[ht]
    \centering
    \caption{List of additional variables and parameters used in our MADDM system.}
     \begin{tabular}{p{0.35cm}p{7cm}}
     \hline
     \multicolumn{2}{c}{Variables and Parameters List} \\
     \hline
     $x$ & advisor index \\
     $d$ & decision index\\
     $s$ & selection function\\
     $f$ & decision function\\
     $\alpha_x$ & correct estimated evidence of the advisor $x$\\
     $\beta_x$ & wrong estimated evidence of the advisor $x$\\
     $\theta_x$ & uncertainty of the advisor $x$\\
     $\tau_x$ & trustworthiness of the advisor $x$\\
     $\tau'_x$ & trustworthiness of the advisor $x$ from Beta Sampling\\
     $i_d$ & confidence value of decision $d$\\
     $c_x$ & price of the advisor $x$\\
     $P_d$ & set of the advisors whose answer for decision $d$ is $1$\\
     $N_d$ & set of the advisors whose answer for decision $d$ is $-1$\\
     $Y_d$ & $P_d\cup N_d$\\
     $u_d$ & utility of decision $d$\\
     $a_d$ & final inferred answer of decision $d$\\
     $a^*_d$ & ground truth of decision $d$\\
     $v^+_d$ & profits that if $a_d = a^*_d$\\
     $v^-_d$ & loss that if $a_d \neq a^*_d$\\
     $P^{e+}_d$ & probability that $a_d = 1$ from ensemble model\\
     $P^{e-}_d$ & probability that $a_d = -1$ from ensemble model\\
     \hline
    \end{tabular}

    \label{tab:variables_and_parameter}
\end{table}
Accordingly, for every decision, $d$, the total cost to the decision-maker to hire the advisors in $s(I)$ is $C_d(s) = \sum_{x \in s(I)}c_{x}$.

Finally, we define the utility that the decision-maker gets for every decision. Given a decision, $d$, we define its utility to the decision-maker as $u_d(s) = v_d - C_d(s)$.
In particular, the sum of the utilities for all the decisions is $u(s) = \sum_{d \in D}u_{d}(s)$. Since each advisor has a different cost, the final utility depends on the advisor selection function adopted. 
In this framework, the goal of the decision-maker is to find the selection function, $s$, that maximizes its payoff:
\begin{equation}
	s^* = \mathop{\arg\max}_{s\in \mathcal{S}}u(s),
\label{eq:utility}
\end{equation}
where $\mathcal{S}$ denotes the set of all feasible selection functions.

\section{Multi-Advisor Dynamic Decision-Making} \label{Sec: MTIRL}
\label{sec:mtirlformal}
\begin{figure*}[ht]
   \centering
   \includegraphics[scale=0.5]{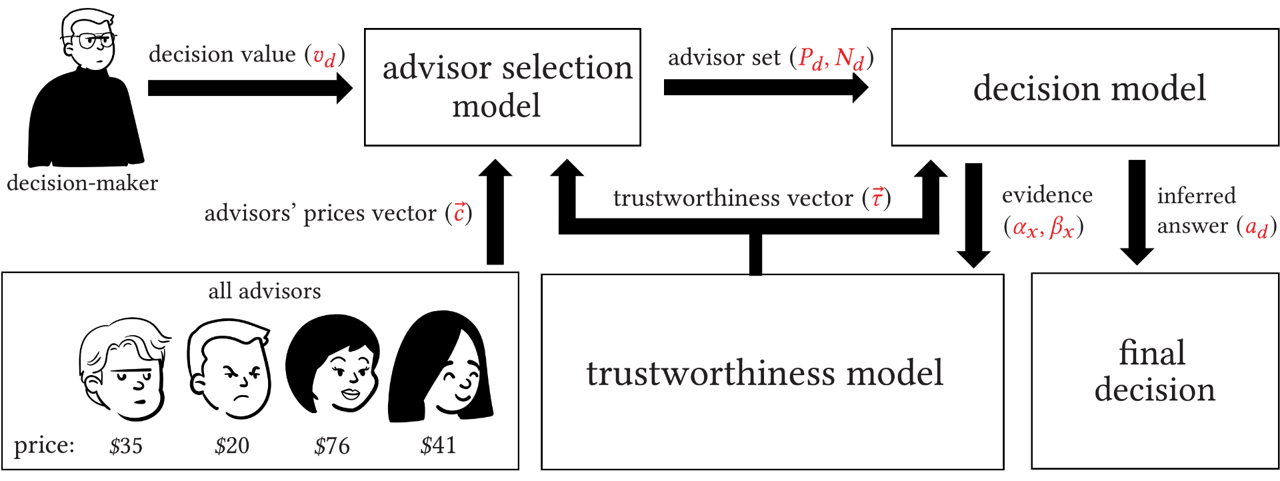}
   \caption{The advisor selection model can select a subset of advisors from all advisors to answer the decision. It needs to consider the decision of value and risk, the advisors' cost, and trustworthiness. The decision model uses advisors' trustworthiness and the answer set to decide the aggregating answer and the estimated evidence for updating the trustworthiness. The trustworthiness model builds and updates the  advisors' trustworthiness.}
   \label{Fig:MADDM}
\end{figure*}

The design of MADDM consists of three components.
The first is a trust assessment model that determines an advisor's trustworthiness, which can be used as a weight in the decision model and to calculate the contributions of advisors in the advisor selection model.  
The second component is the advisor selection model, which assigns a set of advisors to every decision.
The third is the decision model, which selects an answer after receiving the advisors' opinions. 
Figure \ref{Fig:MADDM} provides a graphical overview of the structure of MADDM. 

\subsection{Trustworthiness Model} \label{sec:trustformal}
Following J{\o}sang \cite{josang2016subjective}, we build our trustworthiness model using a Beta distribution. 
Recall that we do not know the ground truth. So, for each advisor, we associate two values, called \textit{advice estimated to be correct} $\alpha_x$ and \textit{advice estimated to be incorrect} $\beta_x$. 
Initially, these values are $1$ \cite{josang2016subjective}; i.e.\ we start with a prior that is $\text{Beta}(1,1)$, or close to uninformative. We update these values whenever the advisor responds to a query. Correct answers to all decisions are \emph{estimated} by our model without ground truth; we use the estimated answer to determine whether the advisor's answer is correct or not (see Section \ref{sec:bawvedm}).

Now, for each advisor $x \in X$, we define its trustworthiness as $\tau_x = \alpha_x/(\beta_x +\alpha_x) \in (0, 1)$. 
If $\tau_x = 1$, we say that the advisor $x$ is completely trustworthy.
If $ \tau_x = 0$, we say that the advisor $x$ is completely untrustworthy.

This concept of trustworthiness is insufficient since it does not capture the epistemic uncertainty associated with that assessment.
For this reason, each advisor's trustworthiness $\tau_x$ is paired with a parameter that quantifies this epistemic uncertainty behind the computation of $\tau_x$.
This uncertainty will reduce as we acquire more evidence regarding an advisor $x$. 
More specifically, we compute the uncertainty by using \emph{Subjective Logic} \cite{josang2016subjective}. This is commonly-employed method in computational models of trust in multi-agent systems and information fusion. 
Formally, for each advisor $x \in X$, the uncertainty of $x$ is $\theta_x = 2/(\alpha_x+\beta_x) \in (0, 1]$. 
% 
%If $\theta_x = 1$, we say that there is no estimated evidence on the trustworthiness of advisor $x$, i.e. $\alpha_x = \beta_x = 1$. 
%
%Notice that, as the values of $\alpha_x$ and $\beta_x$ increase, the uncertainty $\theta_x$ monotonically decreases from $1$ and tends to $0$.  

\subsection{Advisor Selection}
\label{sec:as}
The overall aim of the system is to maximize utility, $u(s)$ (see Equation \ref{eq:utility}), which requires balancing the trade-off between advisor costs and decision value. Typically, the costs of asking all advisers would exceed the decision value, even if the decision is correct, so it is rarely optimal.  For example, for a decision with a value of $\$$10, it is not worth spending $\$$100 to hire advisors. 

Our method selects the set of advisors according to the value of the problem and estimates their contributions to a decision. We assume their trustworthiness is initially unknown, and all advisers have equal trustworthiness. This knowledge is updated over time but is not reliable at first. Therefore, focusing too early on seemingly good advisors can lead to sub-optimal decisions.
To address this, our system solves a multi-armed bandit problem in which it has to balance the exploration of new advisors with the exploitation of the knowledge it has already gathered.
Among the many possible algorithms used to solve the multi-armed bandit problem, we use \textit{Thompson Sampling} \cite{agrawal2012analysis}, which samples from a Beta distribution to compute the contribution of each advisor.
In Algorithm \ref{alg:asa}, we sketch the pseudo-code of our selection function $s$. Recall that $\vec\tau$ denotes the trustworthiness vector that contains the trustworthiness of each advisor, and $\vec c$ are their costs. Let $\vec\alpha$ and $\vec\beta$ denote the estimated evidence vectors, respectively.
Given a decision $d\in D$, let $P^{e+}_d$ and $P^{e-}_d$ denote the probability that $a_d=1$ and $a_d=-1$, respectively.\footnote{These values are computed by the decision model, as we will see in Section \ref{sec:bawvedm}.}
We denote with $U_d$ the vector containing the advisors' utilities. 

\begin{algorithm}[htb]  
  \caption{Pseudo-code of the Advisor Selection algorithm}  
  \label{alg:asa}  
  \begin{algorithmic}[1]
    \State  \textbf{Input}: $d, \vec\tau, \vec\alpha, \vec\beta, v^+_d, v^-_d, \vec c$
    \State $initialize$ $P^{e+}_d = P^{e-}_d = 0.5, U_d = P_d = N_d = \varnothing$
    \State \textbf{while} $true$ \textbf{do}
        \State \qquad \textbf{for} advisor $x$ in $X$ \textbf{do}
        \State \qquad \qquad $\tau'_x \leftarrow ThompsonSampling(\alpha_x, \beta_x)$
        \State \qquad \qquad $u_{d,x} \leftarrow UtilityComputation(\tau'_x, v^+_d, v^-_d, P^{e+}_d, P^{e-}_d)$
        \State \qquad \qquad $U_d.append(u_{d,x})$    
        \State \qquad $u_{d,x^*} = Max(U_d)$
        \State \qquad \textbf{if} $u_{d,x^*} > 0$ \textbf{do}
        \State \qquad \qquad \textbf{if} $a^{x^*}_d = 1$ \textbf{do}
        \State \qquad \qquad \qquad $P_d.append(x^*)$
        \State \qquad \qquad \textbf{if} $a^{x^*}_d = -1$ \textbf{do}
        \State \qquad \qquad \qquad $N_d.append(x^*)$
        \State \qquad $P^{e+}_d, P^{e-}_d \leftarrow DecisionModel(P_d, N_d, \vec\tau)$
        \State \qquad $U_d = \varnothing$ 
        \State \qquad $X.remove(x^*)$
    \State \textbf{until} $u_{d,x^*} \leq 0$
    \State  \textbf{Output}: $P_d, N_d$
  \end{algorithmic}  
\end{algorithm} 

In more detail, after initializing the answer probabilities $P^{e+}_d$ and $P^{e-}_d$, the answer sets $P_d$ and $N_d$, the utility vector $U_d$, and the trustworthiness vector (Line 2), the model enters a loop for selecting advisors (Line 3).
Let $V^x_d$, $u^x_d$ denote the expected contribution and the marginal utility of the advisor $x$ in decision $d$. Recall that $c_x$ is the price of advisor $x$. Their relationship can be expressed as follows:
  \begin{equation}
  \label{Equ:uqx}
    u^x_d = V^x_d - c_x.
  \end{equation}
In each round of advisor selection, we need to compute the marginal utility $u^x_d$ of each advisor and select the one with the best $u_{d,x^*}$, which is our estimate of the advisor that maximizes the expected profit for the decision-maker (Lines 4-8).  

Computing the marginal utility $u^x_d$ is achieved in two steps. First, for each advisor, $x$, we define a Beta distribution $\text{Beta}(\alpha_x, \beta_x)$ and sample from it to get the Beta trustworthiness $\tau'_x$. We only use it to compute the utility $u_{d,x}$ of the advisor $x$ (Line 5), whereas the model does not use $\tau'_x$ for real decision-making. When there is little evidence regarding an advisor, e.g.\ when $\alpha = 1$ and $\beta = 1$, the Beta distribution has a large variance. Consequently, the value $\tau'_x$ is subject to large fluctuations, which increases the decision error. 

Second, we need to know the contribution $V^x_d$ of each advisor $x$.  Let us now assume that advisor $x$ answered $1$ to a decision $d$; the case in which the advisor answers $-1$ follows a similar routine. In order to compute its contribution, we first add $x$ to the set $P_d$ and proceed to calculate the probabilities $P^{e+'}_d$ and $P^{e-'}_d$. The value $P^{e+'}_d$ and $P^{e-'}_d$ describes the probability that $a_d = 1$ and $a_d = -1$, respectively. Therefore, the wider the gap between $P^{e+}_d$ and $P'_{e+}$, the larger the advisor's contribution. We therefore set:

  \begin{equation}
  \label{Equ:vxq1}
    \Delta V^x_{d,+} = P_+|P^{e+'}_d- P^{e+}_d|*(v^+_d+v^-_d).
  \end{equation}
The values $P_+:= P(a_d^* = 1)$ and $P_-:= P(a_d^* = -1)$ are the \textit{a priori} probability that the answer is positive or negative, respectively.
Hence the value $|P^{e+'}_d - P^{e+}_d|$ represents the change of the answer probability if advisor $x$ participates in the decision. Similarly, if the advisor answers $-1$, we set:
  \begin{equation}
  \label{Equ:vxq2}
    \Delta V^x_{d,-} = P_-|P^{e-'}_d- P^{e-}_d|*(v^+_d+v^-_d).
  \end{equation}

After we compute $\Delta V^x_{d,+}$ and $\Delta V^x_{d,-}$, we compute the expected contribution $V^x_d$ as:
  \begin{equation}
  \label{Equ:caq}
    V^x_d = (\tau'_x - (1-\tau'_x)) * (\Delta V^x_{d,+} + \Delta V^x_{d,-}).
  \end{equation}
Finally, the algorithm computes the utility $u^x_d$ by Equation \ref{Equ:uqx}.

If $u^{x^*}_d > 0$, the advisor $x^*$ is selected, which means that its contribution is greater than its cost. The selected advisor $x^*$ needs to provide the answer for decision $d$. Depending on the answer from the advisor $x^*$, it can be added to $P_d$ or $N_d$ (Lines 9--13), which is used to update the answer probability $P^{e+}_d$ and $P^{e-}_d$ (Line 14). After every selection, we need to recalculate the marginal utility of each advisor for selecting the next advisor because their marginal utilities change. For example, if we select an advisor with $90\%$ trustworthiness and give a positive answer to decision $d$, $P^{e+}_d$ will increase from $50\%$ to $90\%$. The model repeats Lines 4--16 to select advisors one by one until $u_{d,x^*} \leq 0$ (Line 17), and outputs the final answer set $(P_d, N_d)$ (Line 18).

\subsection{Bayesian and Weighted Voting Ensemble Decision Model} \label{sec:bawvedm}
We use Bayesian and Weighted Voting Ensemble (BWVE) as the decision function $f$ to make decisions \cite{zhaori2022}. There are two reasons for choosing BWVE. First, it is a truth inference method without ground truth. It has been shown to outperform the simple weighted voting method, which considers the advisors' weights, determined by their trustworthiness, to bias majority voting \cite{zhaori2022}. Second, it returns a probability distribution over the answers, allowing us to evaluate each advisor's contribution, which aligns with our advisor selection model and retrospectively re-calibrates their trustworthiness.

In the following, we detail the BWVE procedure. 
Essentially, it combines two decision procedures to improve the overall outcome. One is based on a Bayesian model, while the other follows a weighted voting decision method. If we know the real trustworthiness $\vec \tau$ of all the advisors, the Bayesian method will obtain higher accuracy than the weighted voting method. However, in the beginning, because the uncertainty of the trustworthiness is large, the Bayesian method is unstable, so BWVE relies more on the weighted voting method for decisions. With the decreasing of the average uncertainty, the Bayesian method has a better performance. So BWVE uses the average uncertainty to control the weights of Bayesian and weighted voting automatically. 
\subsubsection{Bayesian} \label{sec:bawvedmbp}
For every decision $d$, the advisor selection function returns a subset $Y_d \subseteq X$ that needs to answer the decision $d$.
We recall that $P_{d} \subseteq Y_d$ denotes the set of advisors that answered $1$ to the decision, while the advisors in $N_{d}  \subseteq Y_d$ answered $-1$.
Given the partition $(P_{d},N_{d})$ of $Y_d$, from the Bayesian method, the probability that $a^*_d= 1$ is $P^{b+}_d:=P_b(a^*_d = 1|P_{d}, N_{d})$, while $P^{b-}_d:=P_b(a^*_d = -1|P_{d}, N_{d})$ is the probability that $a^*_d=-1$.
From Bayes theorem, we can then express $P^{b+}_d$ and $P^{b-}_d$ as follows:

%   \begin{equation}
  \begin{align}\label{Equ:P(fpc)}
    P^{b+}_d =\frac{P_{+}P(P_{d},N_{d}|a^*_d = 1)}{P_{+}P(P_{d}, N_{d}|a^*_d=1)+ P_{-}P(P_{d},N_{d}|a^*_d=-1)}     
  \end{align}
  
%   \end{equation}
  \begin{equation}
  \label{Equ:P(fnc)}
    P^{b-}_d = \frac{P_{-}P(P_{d},N_{d}|a^*_d = -1)}{P_{-}P(P_{d}, N_{d}|a^*_d=-1)+P_{+}P(P_{d},N_{d}|a^*_d=1)}.
  \end{equation}
We recall that $P_+:= P(a_d^* = 1)$ and $P_-:= P(a_d^* = -1)$ is the \textit{a priori} probability that the answer is positive or negative, respectively.
Since we do not have any evidence about $a^*_d$, both $P_{+}$ and $P_{-}$ are equally likely, therefore we set $P_{+}=P_{-}=0.5$.
The quantities $P(P_{d}, N_{d}|a^*_d=1)$ and $P(P_{d}, N_{d}|a^*_d=-1 )$ describe the probability to observe the partition $(P_{d}, N_{d})$ under the assumption that $a^*_d=1$ and $a^*_d=-1$, respectively.
Both $P(P_{d}, N_{d}|a^*_d=1)$ and $P(P_{d}, N_{d}|a^*_d=-1 )$ are computed through the trustworthiness $\tau_x$ as it follows:
  \begin{equation}
  \label{Equ:P(cfp)}
    P(P_{d}, N_{d}|a^*_d=1) =\prod_{i \in P_d} \prod_{j \in N_d} \tau_i(1-\tau_j)
  \end{equation}
    \begin{equation}
  \label{Equ:P(cfn)} 
    P(P_{d}, N_{d}|a^*_d=-1) =\prod_{i \in P_d} \prod_{j \in N_d} \tau_j(1-\tau_i)
  \end{equation}
\subsubsection{Weighted Voting} \label{sec:bawvedmwvti}
The Bayesian decision method can only work well when the advisors' trustworthiness is sufficiently high. In the initial phase of the process, the advisors' trustworthiness is unreliable, so the Bayesian method is not stable. Since there is no ground truth, it is easily misled by bad advisors when the mean advisors' accuracy is not high. BWVE deals with this problem by using the weighted voting method, which is more robust than Bayesian at the beginning. Then, during the initialization, the weighted voting method has more influence on the decision than Bayesian. For the weighted voting method, under the answer set ($P_{d}$, $N_{d}$), the probability that the ground truth $a^*_d$ = 1 and -1 are correct can be denoted as $P^{w+}_d:=P_{wv}(a^*_d = 1|P_{d}, N_{d})$ and $P^{w-}_d:=P_{wv}(a^*_d = -1|P_{d}, N_{d})$, respectively. The model then uses the sum of the advisors' trustworthiness to calculate them:
  \begin{equation}
  \label{Equ:P(fpcv)}
 P^{w+}_d = \frac{\sum_{i \in P_d}\tau_i}{\sum_{j \in P_d \cup N_d}\tau_j}
  \end{equation}
   \begin{equation}
  \label{Equ:P(fncv)}
 P^{w-}_d = \frac{\sum_{i \in N_d}\tau_j}{\sum_{j \in P_d \cup N_d}\tau_j}
  \end{equation}
  
\subsubsection{Ensemble Decision}
BWVE uses the average uncertainty $\bar{\theta}_{d}$ to control the weights of the Bayesian and the weighted voting for decisions.
The higher the average uncertainty of the advisors in the answer set $Y_d$, the lower reliability of trustworthiness and the more weight for the weighted voting method. Let $|Y_d|$ denote the cardinality of $Y_d$. It can be expressed as:
   \begin{equation}
  \label{Equ:ut}
  \bar{\theta}_{d}=\frac{\sum_{i \in Y_d}\theta_{i}}{|Y_d|}
  \end{equation}
The average uncertainty $\bar{\theta}_{d}$ gradually decreases as time goes on, and the weight of the Bayesian method needs to increase. For the ensemble decision, given the answer set $(P_{d},N_{d})$, the probability that $a^*_d= 1$ is $P^{e+}_d:=P_b(a^*_d = 1|P_{d}, N_{d})$, while $P^{e-}_d:=P_b(a^*_d = -1|P_{d}, N_{d})$is the probability that $a^*_d=-1$.
They can be expressed as:
% \footnotesize
  \begin{equation}
  \label{Equ:P(fpfinal)}
   P^{e+}_d = (1-\bar{\theta}_{d})P^{b+}_d + \bar{\theta}_{d}P^{w+}_d
  \end{equation}
   \begin{equation}
  \label{Equ:P(fnfinal)}
   P^{e-}_d = (1-\bar{\theta}_{d})P^{b-}_d + \bar{\theta}_{d}P^{w-}_d
  \end{equation}
% \normalsize
Their relationship is:
   \begin{equation}
  \label{Equ:P(pnre)}
P^{e+}_d + P^{e-}_d = 1
  \end{equation}
After getting $P^{e+}_d$ and $P^{e-}_d$, the system needs to compare them. If $P^{e+}_d > P^{e-}_d$, the final answer $a_d = 1$. Otherwise, $a_d = -1$.

\subsubsection{Trustworthiness Update} \label{sec:dmute}
BWVE uses the absolute difference of $P^{e+}_d$ and $P^{e-}_d$ as the new estimated evidence to update $\alpha$ and $\beta$.
  \begin{equation}
  \label{Equ:P(itcon)}
 i_d =  |P_{e}(a^*_d = 1|P_{d}, N_{d}) - P_{e}(a^*_d = -1|P_{d}, N_{d})|
  \end{equation}
If $a_d = 1$, the update of $\alpha_{x}$ and $\beta_{x}$ can be expressed as:
   \begin{equation}
  \label{Equ:pabu}
\begin{array}{l}
     \alpha_{x}\leftarrow \alpha_{x}+i_{d} \quad\forall x \in P_{d},\\\\
     \beta_{x}\leftarrow\beta_{x}+i_{d}\quad\forall x \in N_{d},
\end{array}
  \end{equation}
If $a_d = -1$, the update of $\alpha_{x}$ and $\beta_{x}$ can be expressed as:
   \begin{equation}
  \label{Equ:nabu}
\begin{array}{l}
     \beta_{x}\leftarrow\beta_{x}+i_{d}\quad \forall x \in P_{d},\\\\
    \alpha_{x}\leftarrow\alpha_{x}+i_{d}\quad\forall x \in N_{d}, 
\end{array}
  \end{equation}
  
\subsubsection{Review Update} \label{sec:RM}
Recall that MADDM is an online problem without access to ground truth. Moreover, the initial trustworthiness is low. Therefore, the update of the trustworthiness $\vec \tau$ relies on the evidence from new decisions. And the decisions, in turn, rely on the trustworthiness $\vec \tau$. This dynamic loop is used for building the model to make the trustworthiness and the aggregating answer more accurate. Therefore, similar to the EM method, after every answer, we continuously update the trustworthiness of the advisors through the answers from past decisions. 
Algorithm \ref{alg:asa2} describes how the review update works. Let $\vec P_{past}$, $\vec N_{past}$ denote the vector that contains the past answer set, and we recall that $\vec\tau$ denote the trustworthiness vector that contains all advisors' trustworthiness. 
Let $\vec\tau_0$ denote the old trustworthiness vector, $\Delta\tau$ the sum of the difference between the old trustworthiness vector $\vec\tau_0$ and the new trustworthiness vector $\vec\tau$. Furthermore, let $V_s$ denote the threshold of $\Delta\tau$ for terminating the update. $V_s$ usually is set to a small value. Note that $\Delta\tau$ is used to judge the update step size of $\vec\tau$. Specifically, when $\Delta\tau$ is smaller than $V_s$, the model stops updating.
  \begin{algorithm}[htb]  
  \caption{Pseudo-code of the review maximization algorithm}  
  \label{alg:asa2}  
  \begin{algorithmic}[1]
    \State  \textbf{Input}: $\vec P_{past}, \vec N_{past}, \vec\tau, V_s$ 
    \State $initialize$ $\Delta\tau=0$, $\vec\tau_0 = \vec\tau$
    \State \textbf{while} $true$ \textbf{do}
        \State \qquad \textbf{for} $P_d, N_d$ in $\vec P_{past}, \vec N_{past}$ \textbf{do}
        \State \qquad \qquad $P^{e+}_d, P^{e-}_d \leftarrow f(P_d, N_d, \vec\tau)$
        \State \qquad \qquad $\vec\tau_0 = \vec\tau$
        \State \qquad \qquad $\vec\tau \leftarrow TrustworthinessUpdate(P^{e+}_d, P^{e-}_d)$
        \State \qquad \qquad $\Delta\tau = sum(\vec\tau - \vec\tau_0)$
    \State \textbf{until} $\Delta\tau \leq V_s$
    \State  \textbf{Output}: $\vec\tau$
  \end{algorithmic}  
\end{algorithm} 

\section{Experiments} \label{Sec: mtexper}
In this section, we present the decision-answer experiments to evaluate our method.
Specifically, we compare our method with two cost-constraint-based methods. The first is the Fixed Number of Advisors based method (FNA), which means that the decision-maker selects a fixed number for answering every decision. The second is the Budget-Constraint based method (BC), which means that there is a budget constraint to stop selecting advisors. For both approaches, we combine these with different advisor-selection criteria.

\subsection{Setting} \label{Sec:setting}

To the best of our knowledge, there is no standard environment to run decision experiments. For this reason, we rely on synthetically generated ones.
In more detail, the environment we generate includes $1000$ decisions with binary answers and different values.
The full set of advisors consists of $30$ simulated agents with different answer accuracy and costs.
An Extended Rectified Gaussian distribution (ERGd) samples both the profits and losses of every decision \cite{palmer2017methods}.
We generate each advisor's real accuracy and cost using the same probability distribution. 
During the experiments, the decision-maker selects a set of advisors to enquire and infers the answers using different methods.
After answering $1000$ decisions, the decision-maker gets the final utility. 
Due to the probabilistic nature of the experiments, every experiment is repeated for $100$ different runs  to obtain statistically significant results.  
To reduce variance and bias, all methods are run using the same conditions. That is, although the conditions vary between runs, the same set of runs are used to compare the methods (i.e., using the same set of advisor qualities and  prices, the same decision sequence, and the same decision profits and losses).

We consider different ratios between the decision's value and the advisor's cost, which leads us to define two sets of experiments. 
In the first set, both the decision profits and decision losses are sampled from an ERGd whose means and standard deviation are equal to $100$.
In the second one, the mean and the standard deviation of the ERGd are both changed to $500$.
Due to the large deviation, the decision values are highly volatile. Hence, some decisions may be worth more than $1000$, and some may be worthless. 

Furthermore, the real accuracy of advisor $x$, i.e. $\tau^r_x$, is sampled from an ERGd whose standard deviation is fixed at $0.3$ while its mean ranges in the set $\{0.5+0.01*k\}$ where $k=1,2,\dots,50$.
For example, if $\tau^r_x$ is $0.8$, the advisor $x$ has $80\%$ probability of giving a correct answer.
Hence, we consider $50$ different frameworks in which the average trustworthiness increases every time.
Finally, we assume that the cost of each advisor is proportional to its real trustworthiness.
In practice, higher quality often comes at a cost. For example, senior advisors are more costly than junior ones. Similarly, more advanced machine learning algorithms typically require higher computational costs. However, this is only a correlation and not always the case for every instance. To achieve this correlation, the cost of each advisor is sampled from an ERGd whose average is $\tau^r_x*20$ and whose standard deviation is $10$. Note that the correlation makes the problem more challenging since the system has to make trade-offs between cost and quality. Without such correlation, there is a high likelihood of a cheap and reliable advisor which makes the problem easier to solve but also less realistic. 

\begin{table}[t]
  \caption{Experiment setting}
  \label{tab:exst}
  \begin{tabular}{rll}\toprule
    setting & value\\ \midrule
    env1: decision profits $v^+_d$ $mean,std$ & $100, 100$ \\
    env1: decision loss $v^-_d$ $mean,std$ & $100,100$  \\
    env2: decision profits $v^+_d$ $mean,std$ & $500,500$ \\
    env2: decision loss $v^-_d$ $mean,std$ & $500,500$ \\
    advisor cost $c_x$ $mean,std$  & $0$ to $20$,$10$ \\
    real trustworthiness $mean,std$ & from $0.51$ to $1$,$0.3$ \\ \bottomrule
  \end{tabular}
\end{table}

We used three different exploration methods (Upper Confidence Bound (UCB), Thompson Sampling, $\epsilon$-greedy) and two rules of the advisor selection (trustworthiness, cost-effectiveness) to combine with FNA and BC, respectively. The aggregation method of FNA and BC is EM, which can maximize the sample utilization and has been verified multiple times in truth inference \cite{demartini2012zencrowd,gemalmaz2021accounting}.

In more detail, in terms of advisor selection strategies, UCB, Thompson Sampling and $\epsilon$-greedy are effective for solving the multi-armed bandit problem. We experimented with a range of values and found that the $\epsilon$-greedy method has the best performance when $\epsilon = 0.1$ (we also tested $\epsilon$ = 0.05, 0.15, 0.2, 0.25) for all methods. UCB and Thompson Sampling explore more than $\epsilon$-greedy at the beginning. Since the lack of ground truth, not every exploration can provide correct feedback for updating trustworthiness, especially when the average advisor's accuracy is low.

\begin{figure*}[h!]
    \centering
    \subfigure[\normalsize environment $1$: mean acc, std = 100, 100]
    {
    \begin{minipage}[t]{1\linewidth}
    
      \centering
      \includegraphics[scale=0.37]{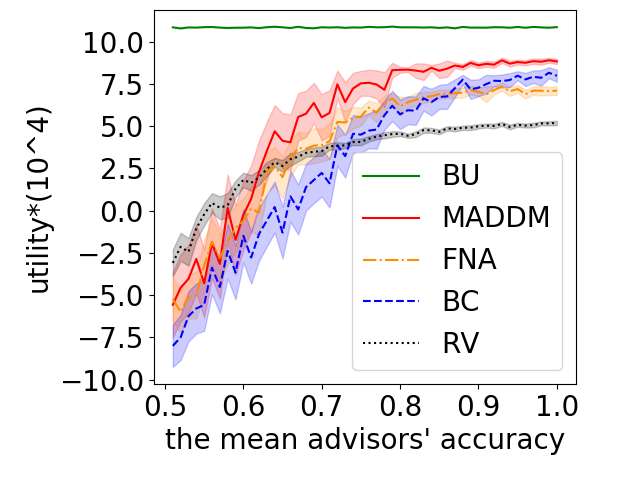}
    %   \vspace{0.01cm}
          \includegraphics[scale=0.37]{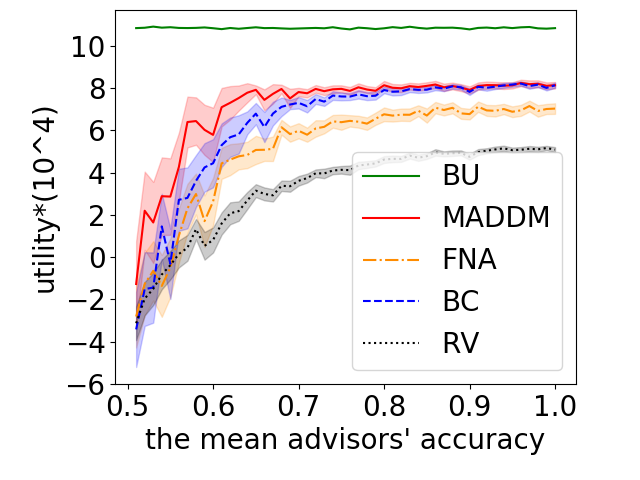}
      \end{minipage}
    }
    \subfigure[\normalsize environment $2$: mean acc, std = 500, 500]
    {
              \begin{minipage}[t]{1\linewidth}
              \centering
          \includegraphics[scale=0.37]{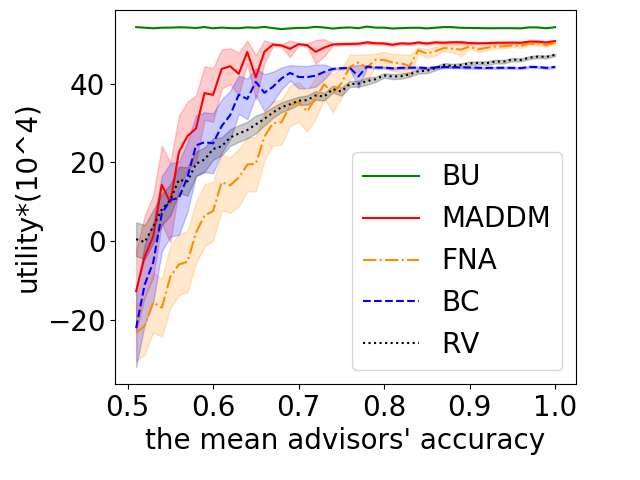}
          \includegraphics[scale=0.37]{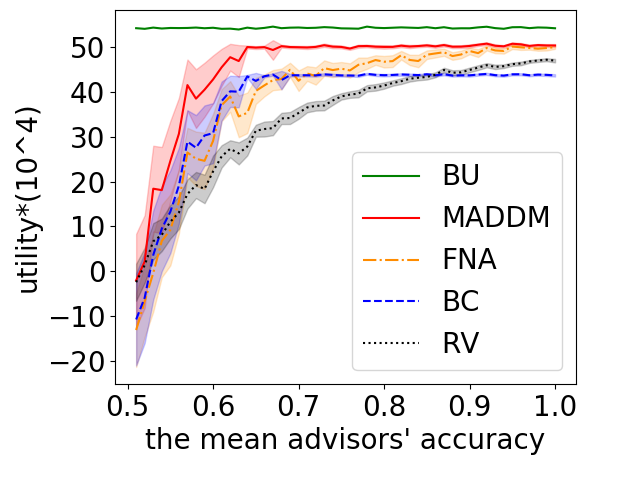}
              \end{minipage}
    }
    \caption{In four figures, the X-axis represents the mean advisors' accuracy from 0.51 to 1. The Y-axis represents the average utility of 100 experiments. The half-transparent area, along with the curve, is the 95\% confidence interval error bar. Figure (a) shows the results of environment 1 (mean, standard deviation = 100, 100). Figure (b) shows the results of environment 2 (mean, standard deviation = 500, 500). In Figures (a) and (b), the left figures represent the standard methods, and the right figures are the exploration-first-based methods. MADDM = multi-advisor dynamic decision-making(ours); FNA = $\epsilon$-greedy fixed number of the advisor EM; BC = $\epsilon$-greedy budget-limited EM; RV = random voting.
     }
    \label{fig:results}
\end{figure*}

The criteria for advisor selection contain trustworthiness and cost-effectiveness. For example, if trustworthiness is the rule, the greedy strategy always selects the advisor with bigger trustworthiness. Cost-effectiveness is a method we improved from work \cite{xia2015thompson}. The cost-effectiveness of the advisor $x$ can be expressed by $c_x/(\tau_x-0.5)$, which means how much cost is the improvement of trustworthiness for advisor $x$. It has a better performance than trustworthiness. 

For FNA and BC, we also test their performance under different hyper-parameters. First, we test the performance of FNA by setting the number of advisors from $1$ to $10$. The results show that five advisors have the best performance. Second, BC, we try $5\%$, $10\%$, $15\%$, $20\%$, $25\%$ of the value (profit + loss) of every decision as the budget constraint and $10\%$ has the best performance. 

To clearly understand the performance of our method, FNA and BC, we selected two other methods for comparison. The first is random voting (RV). It randomly selects three advisors and combines them by majority voting. Another one is the best utility (BU). It describes the maximum utility the decision-maker can get, which means all the decisions are correct, and the advisor cost is $0$. 

The method with the trustworthiness model is easily misled by malicious advisors when the mean advisors' accuracy is low \citep{zhaori2022}. In practical applications, the methods for solving the problem include adding some decisions with ground truth, selecting several advisors with high accuracy to participate in decision-making, or considering the prior information of advisors. In this paper, our assumptions are no ground truth and no prior information, so we design the exploration-first model to solve this problem. In the first few decisions, the model selects all advisors for answering to increase the accuracy of the answer and then back to the method's standard advisor selection strategy. We use this model before rounds $1$-$15$, respectively, and the results show that the three methods perform best when the model is used before the $10$ round. Therefore, we added the exploration-first model to our method, FNA and BC and did additional experiments in two environments.

\subsection{Results} \label{Sec:results}
We now compare the utility obtained by the different methods we considered. 
Table \ref{tab:avre} shows the mean and standard deviation of the utility in every environment. 
Overall, our MADDM method has the best performance in terms of the average utility in almost all environments.
In all the experiments, the average utilities obtained by the exploration-first methods are significantly bigger than the others. 
Moreover, the standard deviation of the utilities is also reduced, which means that the result is more consistent. 
We did $600$ (3*50*4) pairs of Mann-Whitney Tests between MADDM and FNA, BC, and Random Voting (RV) with $50$ different average advisors' accuracy in four different environments. 
We observe that $527$ out of the $600$ results have significant differences ($p<0.05/3$).

\begin{table}[t]
  \caption{The red colour numbers The meaning of abbreviations are: env1(SD): environment $1$ and standard methods, env2(SD): environment $2$ and standard methods env1(EF): environment $1$ and all methods with exploration-first model, env2(EF): environment $1$ and all methods with exploration-first model.}
  \label{tab:avre}
  \begin{tabular}{p{1cm}p{1.4cm}p{1.4cm}p{1.4cm}p{1.4cm}}\toprule
     & MADDM & FNA & BC & RV \\ \midrule
    env1(SD)& $\textcolor[rgb]{ 1,  0,  0}{5.19\newline{}\pm 7.68} $  & $3.76\pm$5.95 & 2.85$\pm$7.51 & 3.27$\pm$2.84 \\
    env2(SD)&$\textcolor[rgb]{ 1,  0,  0}{42.69\newline{}\pm 27.38}$ & 30.89$\pm$33.03 & 35.46$\pm$28.11 & 34.16$\pm$16.31 \\
    env1(EF)&$\textcolor[rgb]{ 1,  0,  0}{7.09\newline{}\pm 4.43}$ & 5.15$\pm$4.32 & 6.26$\pm$4.79 & 3.24$\pm$2.93  \\
    env2(EF)& $\textcolor[rgb]{ 1,  0,  0}{44.97\newline{}\pm 22.84}$ & 38.89$\pm$24.11 & 37.58$\pm$22.68 & 34.14$\pm$16.20  \\
\bottomrule
  \end{tabular}
\end{table}

Figure \ref{fig:results} describes the utility curves of different methods as the advisors' accuracy increases.
In the vast majority of cases, MADDM gets more utility than FNA and BC for all the possible accuracy.
In the right graph in Figures \ref{fig:results}a and \ref{fig:results}b, we compare the utilities when all the methods use the exploration-first based model.
RV is better than the other three methods when the mean advisors' accuracy is low.
When there is no ground truth and a significant proportion of bias, the methods with the trustworthiness model are easily misled by malicious advisors.
Once the trustworthiness model is misled, then malicious users take the initiative to sabotage future decisions. However, we observe that MADDM is less prone to be sabotaged.
This is due to the fact that MADDM selects more advisors at the beginning and decreases as trustworthiness is updated, which means MADDM has stronger robustness to malicious advisors than FNA and BC. 
Similarly, we observe that the performances of MADDM are more robust to the manipulation of the bad advisors when the average cost of the advisors and the decision values are bigger.
Since the decisions in the environment, $2$ are more valuable than the ones in the environment $1$, MADDM chooses more advisors to make decisions together at the beginning in environment $2$, which helps to increase the reliability of the answer. 
Therefore, based on this idea, we partially addressed this issue by using the exploration-first-based methods.
The disadvantage of the exploration-first is that it uses more cost for building trustworthiness.
It does not perform as well as standard methods when the mean advisors' accuracy is high. 
However, we do not know the real distribution of the mean advisors' accuracy and decision values before asking, so it is worth using some cost at first to improve the method's expected utility.

MADDM automatically selects the advisors by balancing the advisor's cost and the decision values without any hyper-parameters, which makes MADDM less prone to select an insufficient number of advisors or to waste costs.
In the two methods based on cost-effectiveness, they need to set the number of advisors and budget proportion to control the advisor cost.
If the prior distribution is unknown, the values of these hyper-parameters are difficult to determine.
Furthermore, if the advisor cost is too small, the reliability of the output answer is not enough.
While if the cost is too high, it causes a waste of advisor costs.
For example, in Figure \ref{fig:results}b, we observe that FNA does not select enough advisors when the mean advisors' accuracy is less than $0.8$, whereas the best performance of BC has a gap with MADDM when the mean advisors' accuracy is higher than $0.65$. 

\section{Conclusion} \label{Sec:conclu}
In this paper, we introduce Multi-Advisor Dynamic Decision-Making Method (MADDM), a novel approach for making optimal decisions in sequential decision-making settings with no ground truth. The model takes into account multiple variables, including the decision of profits and loss, advisors' costs, and trustworthiness. It selects advisors by balancing the advisors' costs and the value of making the correct decisions. It also makes decisions by combining the advice from multiple advisors without access to the ground truth and dynamically learns the trustworthiness of advisors without prior information. We test our method through decision-answer experiments in a simulated environment. We also introduce two benchmark methods, one using a fixed number of advisors (FNA) and another one using a fixed budget (BC), which are combined with state-of-the-art sampling and aggregating methods.  The results show that  MADDM significantly outperforms the benchmark methods.

An interesting direction for future work is moving from binary answers to multiple answers, making our approach applicable to more scenarios. This requires changing the calculations of the probabilities to deal with more than two outcomes. The first challenge in doing so is calculating the confidence value and how to use it for updating the trustworthiness. The second challenge is adjusting the weights of the weighted voting approach and the Bayesian method for making the decision. Another interesting direction is dealing with multiple simultaneous decisions at each point, which requires us to consider the allocation of advisors to each of the decisions. 
\bibliographystyle{ACM-Reference-Format} 
\bibliography{MADDM}


\begin{thebibliography}{20}

%%% the Web address.

\ifx \showCODEN    \undefined \def \showCODEN     #1{\unskip}     \fi
\ifx \showDOI      \undefined \def \showDOI       #1{#1}\fi
\ifx \showISBNx    \undefined \def \showISBNx     #1{\unskip}     \fi
\ifx \showISBNxiii \undefined \def \showISBNxiii  #1{\unskip}     \fi
\ifx \showISSN     \undefined \def \showISSN      #1{\unskip}     \fi
\ifx \showLCCN     \undefined \def \showLCCN      #1{\unskip}     \fi
\ifx \shownote     \undefined \def \shownote      #1{#1}          \fi
\ifx \showarticletitle \undefined \def \showarticletitle #1{#1}   \fi
\ifx \showURL      \undefined \def \showURL       {\relax}        \fi
% invisible to TeX
\providecommand\bibfield[2]{#2}
\providecommand\bibinfo[2]{#2}
\providecommand\natexlab[1]{#1}
\providecommand\showeprint[2][]{arXiv:#2}

\bibitem[\protect\citeauthoryear{Agrawal and Goyal}{Agrawal and Goyal}{2012}]%
        {agrawal2012analysis}
\bibfield{author}{\bibinfo{person}{Shipra Agrawal} {and} \bibinfo{person}{Navin
  Goyal}.} \bibinfo{year}{2012}\natexlab{}.
\newblock \showarticletitle{Analysis of thompson sampling for the multi-armed
  bandit problem}. In \bibinfo{booktitle}{\emph{Conference on learning
  theory}}. JMLR Workshop and Conference Proceedings,
  \bibinfo{publisher}{PMLR}, \bibinfo{address}{Edinburgh, Scotland},
  \bibinfo{pages}{39--1}.
\newblock


\bibitem[\protect\citeauthoryear{Cayci, Eryilmaz, and Srikant}{Cayci
  et~al\mbox{.}}{2020}]%
        {cayci2020budget}
\bibfield{author}{\bibinfo{person}{Semih Cayci}, \bibinfo{person}{Atilla
  Eryilmaz}, {and} \bibinfo{person}{Rayadurgam Srikant}.}
  \bibinfo{year}{2020}\natexlab{}.
\newblock \showarticletitle{Budget-constrained bandits over general cost and
  reward distributions}. In \bibinfo{booktitle}{\emph{International Conference
  on Artificial Intelligence and Statistics}}. PMLR, \bibinfo{publisher}{PMLR},
  \bibinfo{pages}{4388--4398}.
\newblock


\bibitem[\protect\citeauthoryear{Demartini, Difallah, and
  Cudr{\'e}-Mauroux}{Demartini et~al\mbox{.}}{2012}]%
        {demartini2012zencrowd}
\bibfield{author}{\bibinfo{person}{Gianluca Demartini},
  \bibinfo{person}{Djellel~Eddine Difallah}, {and} \bibinfo{person}{Philippe
  Cudr{\'e}-Mauroux}.} \bibinfo{year}{2012}\natexlab{}.
\newblock \showarticletitle{Zencrowd: leveraging probabilistic reasoning and
  crowdsourcing techniques for large-scale entity linking}. In
  \bibinfo{booktitle}{\emph{Proceedings of the 21st international conference on
  World Wide Web}}. \bibinfo{pages}{469--478}.
\newblock


\bibitem[\protect\citeauthoryear{Gemalmaz and Yin}{Gemalmaz and Yin}{2021}]%
        {gemalmaz2021accounting}
\bibfield{author}{\bibinfo{person}{Meric~Altug Gemalmaz} {and}
  \bibinfo{person}{Ming Yin}.} \bibinfo{year}{2021}\natexlab{}.
\newblock \showarticletitle{Accounting for Confirmation Bias in Crowdsourced
  Label Aggregation.}. In \bibinfo{booktitle}{\emph{IJCAI}}.
  \bibinfo{pages}{1729--1735}.
\newblock


\bibitem[\protect\citeauthoryear{Guo, Norman, and Gerding}{Guo
  et~al\mbox{.}}{2022}]%
        {zhaori2022}
\bibfield{author}{\bibinfo{person}{Zhaori Guo}, \bibinfo{person}{Timothy
  Norman}, {and} \bibinfo{person}{Enrico Gerding}.}
  \bibinfo{year}{2022}\natexlab{}.
\newblock \showarticletitle{MTIRL: Multi-trainer interactive reinforcement
  learning system}. In \bibinfo{booktitle}{\emph{International Conference on
  Principles and Practice of Multi-Agent Systems}}. Springer.
\newblock


\bibitem[\protect\citeauthoryear{J{\o}sang}{J{\o}sang}{2016a}]%
        {josang2016generalising}
\bibfield{author}{\bibinfo{person}{Audun J{\o}sang}.}
  \bibinfo{year}{2016}\natexlab{a}.
\newblock \showarticletitle{Generalising Bayes' theorem in subjective logic.}.
  In \bibinfo{booktitle}{\emph{MFI}}. \bibinfo{pages}{462--469}.
\newblock


\bibitem[\protect\citeauthoryear{J{\o}sang}{J{\o}sang}{2016b}]%
        {josang2016subjective}
\bibfield{author}{\bibinfo{person}{Audun J{\o}sang}.}
  \bibinfo{year}{2016}\natexlab{b}.
\newblock \bibinfo{booktitle}{\emph{Subjective logic}}.
  Vol.~\bibinfo{volume}{3}.
\newblock \bibinfo{publisher}{Springer}.
\newblock


\bibitem[\protect\citeauthoryear{Kurenkov, Mandlekar, Martin-Martin, Savarese,
  and Garg}{Kurenkov et~al\mbox{.}}{2020}]%
        {kurenkov2019ac}
\bibfield{author}{\bibinfo{person}{Andrey Kurenkov}, \bibinfo{person}{Ajay
  Mandlekar}, \bibinfo{person}{Roberto Martin-Martin}, \bibinfo{person}{Silvio
  Savarese}, {and} \bibinfo{person}{Animesh Garg}.}
  \bibinfo{year}{2020}\natexlab{}.
\newblock \showarticletitle{AC-Teach: A Bayesian Actor-Critic Method for Policy
  Learning with an Ensemble of Suboptimal Teachers}. In
  \bibinfo{booktitle}{\emph{Proceedings of the Conference on Robot Learning}}
  \emph{(\bibinfo{series}{Proceedings of Machine Learning Research},
  Vol.~\bibinfo{volume}{100})}. \bibinfo{publisher}{PMLR},
  \bibinfo{pages}{717--734}.
\newblock


\bibitem[\protect\citeauthoryear{Landemore}{Landemore}{2012}]%
        {landemore2012collective}
\bibfield{author}{\bibinfo{person}{H{\'e}l{\`e}ne Landemore}.}
  \bibinfo{year}{2012}\natexlab{}.
\newblock \showarticletitle{Collective wisdom: Old and new}.
\newblock \bibinfo{journal}{\emph{Collective wisdom: Principles and
  mechanisms}} (\bibinfo{year}{2012}), \bibinfo{pages}{1--20}.
\newblock


\bibitem[\protect\citeauthoryear{Ly, Marsman, Verhagen, Grasman, and
  Wagenmakers}{Ly et~al\mbox{.}}{2017}]%
        {ly2017tutorial}
\bibfield{author}{\bibinfo{person}{Alexander Ly}, \bibinfo{person}{Maarten
  Marsman}, \bibinfo{person}{Josine Verhagen}, \bibinfo{person}{Raoul~PPP
  Grasman}, {and} \bibinfo{person}{Eric-Jan Wagenmakers}.}
  \bibinfo{year}{2017}\natexlab{}.
\newblock \showarticletitle{A tutorial on Fisher information}.
\newblock \bibinfo{journal}{\emph{Journal of Mathematical Psychology}}
  \bibinfo{volume}{80} (\bibinfo{year}{2017}), \bibinfo{pages}{40--55}.
\newblock


\bibitem[\protect\citeauthoryear{Miao, Peng, Gao, Zhang, and Yin}{Miao
  et~al\mbox{.}}{2022}]%
        {miao2022dynamically}
\bibfield{author}{\bibinfo{person}{Xiaoye Miao}, \bibinfo{person}{Huanhuan
  Peng}, \bibinfo{person}{Yunjun Gao}, \bibinfo{person}{Zongfu Zhang}, {and}
  \bibinfo{person}{Jianwei Yin}.} \bibinfo{year}{2022}\natexlab{}.
\newblock \showarticletitle{On Dynamically Pricing Crowdsourcing Tasks}.
\newblock \bibinfo{journal}{\emph{ACM Transactions on Knowledge Discovery from
  Data (TKDD)}} (\bibinfo{year}{2022}).
\newblock


\bibitem[\protect\citeauthoryear{Palmer, Hill, and Scheding}{Palmer
  et~al\mbox{.}}{2017}]%
        {palmer2017methods}
\bibfield{author}{\bibinfo{person}{Andrew~W Palmer}, \bibinfo{person}{Andrew~J
  Hill}, {and} \bibinfo{person}{Steven~J Scheding}.}
  \bibinfo{year}{2017}\natexlab{}.
\newblock \showarticletitle{Methods for Stochastic Collection and Replenishment
  (SCAR) optimisation for persistent autonomy}.
\newblock \bibinfo{journal}{\emph{Robotics and Autonomous Systems}}
  \bibinfo{volume}{87} (\bibinfo{year}{2017}), \bibinfo{pages}{51--65}.
\newblock


\bibitem[\protect\citeauthoryear{Tao, Jiang, and Li}{Tao et~al\mbox{.}}{2021}]%
        {tao2021differential}
\bibfield{author}{\bibinfo{person}{Fangna Tao}, \bibinfo{person}{Liangxiao
  Jiang}, {and} \bibinfo{person}{Chaoqun Li}.} \bibinfo{year}{2021}\natexlab{}.
\newblock \showarticletitle{Differential evolution-based weighted soft majority
  voting for crowdsourcing}.
\newblock \bibinfo{journal}{\emph{Engineering Applications of Artificial
  Intelligence}}  \bibinfo{volume}{106} (\bibinfo{year}{2021}),
  \bibinfo{pages}{104474}.
\newblock


\bibitem[\protect\citeauthoryear{Tong, Wang, Zhou, Chen, Du, and Ye}{Tong
  et~al\mbox{.}}{2018}]%
        {tong2018dynamic}
\bibfield{author}{\bibinfo{person}{Yongxin Tong}, \bibinfo{person}{Libin Wang},
  \bibinfo{person}{Zimu Zhou}, \bibinfo{person}{Lei Chen},
  \bibinfo{person}{Bowen Du}, {and} \bibinfo{person}{Jieping Ye}.}
  \bibinfo{year}{2018}\natexlab{}.
\newblock \showarticletitle{Dynamic pricing in spatial crowdsourcing: A
  matching-based approach}. In \bibinfo{booktitle}{\emph{Proceedings of the
  2018 International Conference on Management of Data}}.
  \bibinfo{pages}{773--788}.
\newblock


\bibitem[\protect\citeauthoryear{Tran-Thanh, Chapman, Rogers, and
  Jennings}{Tran-Thanh et~al\mbox{.}}{2012}]%
        {tran2012knapsack}
\bibfield{author}{\bibinfo{person}{Long Tran-Thanh}, \bibinfo{person}{Archie
  Chapman}, \bibinfo{person}{Alex Rogers}, {and} \bibinfo{person}{Nicholas
  Jennings}.} \bibinfo{year}{2012}\natexlab{}.
\newblock \showarticletitle{Knapsack based optimal policies for budget--limited
  multi--armed bandits}. In \bibinfo{booktitle}{\emph{Proceedings of the AAAI
  Conference on Artificial Intelligence}}, Vol.~\bibinfo{volume}{26}.
  \bibinfo{pages}{1134--1140}.
\newblock


\bibitem[\protect\citeauthoryear{Tran-Thanh, Stein, Rogers, and
  Jennings}{Tran-Thanh et~al\mbox{.}}{2014}]%
        {tran2014efficient}
\bibfield{author}{\bibinfo{person}{Long Tran-Thanh}, \bibinfo{person}{Sebastian
  Stein}, \bibinfo{person}{Alex Rogers}, {and} \bibinfo{person}{Nicholas~R
  Jennings}.} \bibinfo{year}{2014}\natexlab{}.
\newblock \showarticletitle{Efficient crowdsourcing of unknown experts using
  bounded multi-armed bandits}.
\newblock \bibinfo{journal}{\emph{Artificial Intelligence}}
  \bibinfo{volume}{214} (\bibinfo{year}{2014}), \bibinfo{pages}{89--111}.
\newblock


\bibitem[\protect\citeauthoryear{Wang, Nguyen, Hoang, Dutkiewicz, and
  Cheng}{Wang et~al\mbox{.}}{2018}]%
        {wang2018real}
\bibfield{author}{\bibinfo{person}{Huiyang Wang}, \bibinfo{person}{Diep~N
  Nguyen}, \bibinfo{person}{Dinh~Thai Hoang}, \bibinfo{person}{Eryk
  Dutkiewicz}, {and} \bibinfo{person}{Qingqing Cheng}.}
  \bibinfo{year}{2018}\natexlab{}.
\newblock \showarticletitle{Real-time crowdsourcing incentive for radio
  environment maps: A dynamic pricing approach}. In
  \bibinfo{booktitle}{\emph{2018 IEEE Global Communications Conference
  (GLOBECOM)}}. IEEE, \bibinfo{pages}{1--6}.
\newblock


\bibitem[\protect\citeauthoryear{Xia, Li, Qin, Yu, and Liu}{Xia
  et~al\mbox{.}}{2015}]%
        {xia2015thompson}
\bibfield{author}{\bibinfo{person}{Yingce Xia}, \bibinfo{person}{Haifang Li},
  \bibinfo{person}{Tao Qin}, \bibinfo{person}{Nenghai Yu}, {and}
  \bibinfo{person}{Tie-Yan Liu}.} \bibinfo{year}{2015}\natexlab{}.
\newblock \showarticletitle{Thompson sampling for budgeted multi-armed
  bandits}. In \bibinfo{booktitle}{\emph{Twenty-Fourth International Joint
  Conference on Artificial Intelligence}}.
\newblock


\bibitem[\protect\citeauthoryear{Zheng, Li, Li, Shan, and Cheng}{Zheng
  et~al\mbox{.}}{2017}]%
        {zheng2017truth}
\bibfield{author}{\bibinfo{person}{Yudian Zheng}, \bibinfo{person}{Guoliang
  Li}, \bibinfo{person}{Yuanbing Li}, \bibinfo{person}{Caihua Shan}, {and}
  \bibinfo{person}{Reynold Cheng}.} \bibinfo{year}{2017}\natexlab{}.
\newblock \showarticletitle{Truth inference in crowdsourcing: Is the problem
  solved?}
\newblock \bibinfo{journal}{\emph{Proceedings of the VLDB Endowment}}
  \bibinfo{volume}{10}, \bibinfo{number}{5} (\bibinfo{year}{2017}),
  \bibinfo{pages}{541--552}.
\newblock


\bibitem[\protect\citeauthoryear{Zhou and Tomlin}{Zhou and Tomlin}{2018}]%
        {zhou2018budget}
\bibfield{author}{\bibinfo{person}{Datong Zhou} {and} \bibinfo{person}{Claire
  Tomlin}.} \bibinfo{year}{2018}\natexlab{}.
\newblock \showarticletitle{Budget-constrained multi-armed bandits with
  multiple plays}. In \bibinfo{booktitle}{\emph{Proceedings of the AAAI
  Conference on Artificial Intelligence}}, Vol.~\bibinfo{volume}{32}.
\newblock


\end{thebibliography}

%%%%%%%%%%%%%%%%%%%%%%%%%%%%%%%%%%%%%%%%%%%%%%%%%%%%%%%%%%%%%%%%%%%%%%%%

\end{document}